\documentclass[letterpaper]{article} % DO NOT CHANGE THIS
\usepackage{aaai25}  % DO NOT CHANGE THIS
\usepackage{times}  % DO NOT CHANGE THIS
\usepackage{helvet}  % DO NOT CHANGE THIS
\usepackage{courier}  % DO NOT CHANGE THIS
\usepackage[hyphens]{url}  % DO NOT CHANGE THIS
\usepackage{graphicx} % DO NOT CHANGE THIS
\usepackage{natbib}  % DO NOT CHANGE THIS AND DO NOT ADD ANY OPTIONS TO IT
\usepackage{caption} % DO NOT CHANGE THIS AND DO NOT ADD ANY OPTIONS TO IT
\usepackage{algorithm}
\usepackage{algorithmic}
\usepackage{amsmath}
\usepackage{booktabs}
\usepackage{multirow}
\usepackage{bbding}
\usepackage{xcolor}
\usepackage{stfloats}
\usepackage[linesnumbered,ruled,vlined,algo2e]{algorithm2e}
\usepackage{}
\usepackage{newfloat}
\usepackage{listings}
\DeclareCaptionStyle{ruled}{labelfont=normalfont,labelsep=colon,strut=off} % DO NOT CHANGE THIS
\floatstyle{ruled}
\newfloat{listing}{tb}{lst}{}
\floatname{listing}{Listing}
\title{Boosting Multimodal Large Language Models with Visual Tokens Withdrawal for Rapid Inference}
\author{
    Zhihang Lin\textsuperscript{\rm 1,\rm 2},
    Mingbao Lin\textsuperscript{\rm 3},
    Luxi Lin\textsuperscript{\rm 1},
    Rongrong Ji\textsuperscript{\rm 1}\thanks{Corresponding author}
}
\affiliations{
    \textsuperscript{\rm 1}Key Laboratory of Multimedia Trusted Perception and Efficient Computing, \\
    Ministry of Education of China, Xiamen University, 361005, P.R. China \\
    \textsuperscript{\rm 2}Shanghai Innovation Institute \\
    \textsuperscript{\rm 3}Skywork AI \\
    lzhedu@foxmail.com, linmb001@outlook.com, lewuluu@gmail.com, rrji@xmu.edu.cn
}

\usepackage{bibentry}
\begin{document}

\maketitle

\begin{abstract}
Multimodal large language models (MLLMs) demand considerable computations for inference due to the extensive parameters and the additional input tokens needed for visual information representation. Herein, we introduce Visual Tokens Withdrawal (VTW), a plug-and-play module to boost MLLMs for rapid inference. Our approach is inspired by two intriguing phenomena we have observed: (1) the attention sink phenomenon that is prevalent in LLMs also persists in MLLMs, suggesting that initial tokens and nearest tokens receive the majority of attention, while middle vision tokens garner minimal attention in deep layers; (2) the presence of information migration, which implies that visual information is transferred to subsequent text tokens within the first few layers of MLLMs. As per our findings, we conclude that vision tokens are unnecessary in the deep layers of MLLMs. Thus, we strategically withdraw them at a certain layer, enabling only text tokens to engage in subsequent layers. To pinpoint the ideal layer for VTW, we initially analyze a limited set of tiny datasets and choose the first layer that meets the Kullback-Leibler divergence criterion. Our VTW approach can cut computational overhead by over 40\% across diverse multimodal tasks while maintaining performance. 
% Our code is released at \url{https://github.com/lzhxmu/VTW}.
\end{abstract}

\begin{links}
    \link{Code}{https://github.com/lzhxmu/VTW}
    % \link{Datasets}{https://aaai.org/example/datasets}
    % \link{Extended version}{https://aaai.org/example/extended-version}
\end{links}

\section{Introduction}\label{sec:inro}
% LLM background -> MLLMs background

In recent years, major progress has been made in generative AI with the development of large language models (LLMs)~\cite{achiam2023gpt,team2023gemini,touvron2023llama}. 
Multimodal large language models (MLLMs)~\cite{2024llava,li2023blip2,liu2024llavanext} combine vision encoders, like CLIP~\cite{2021CLIP}, to extract visual features, enhancing LLM's reasoning abilities for complex tasks like visual question answering (VQA)~\cite{lu2022sqa,kembhavi2016ai2d} and visual reasoning~\cite{yue2023mmmu,Liu2023mmbench,fu2023mme}.

However, MLLMs entail high inference cost due to their billions of parameters and their  computational cost quadratic increases with the length of the input sequence.
Converting a high-resolution image into vision tokens further increases the inference cost~\cite{liu2024llavanext}.
The high inference cost of MLLMs hinders their applicability in real-time scenarios.

\begin{figure}[t]
    \centering
    \includegraphics[width=\linewidth]{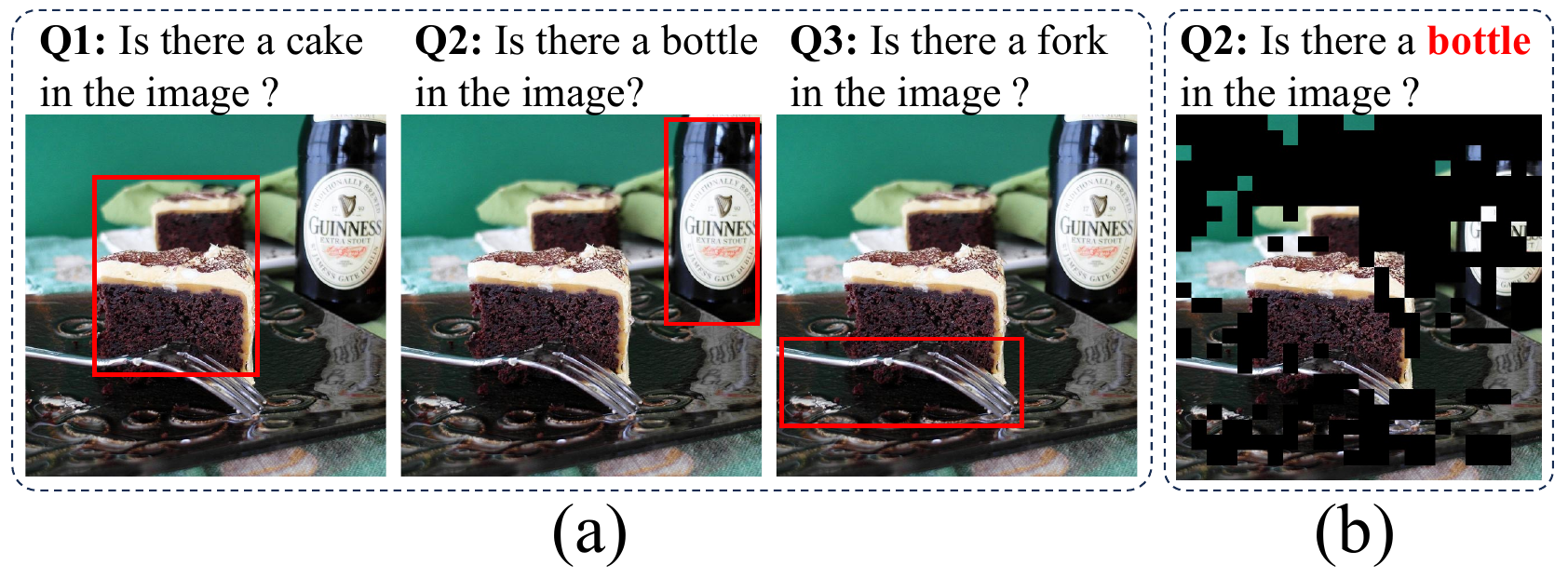}
    % \vspace{-1.0em}
    \caption{
    (a) An instance from POPE~\cite{pope2023kv-cache} with red box indicating key area for answering the question.
    (b) In FastV, some genuinely important tokens like ``bottle'' are pruned, while unimportant tokens like ``cake'' are preserved.
    }
    \vspace{-1em}
    \label{fig:pope_example}
\end{figure}

Recent methods~\cite{shang2024llava-prumerge,chen2024fastv} aim to reduce the computational cost of MLLMs through token reduction.
The primary goal is to keep so-called ``important'' tokens based on predefined metrics and remove or merge the rest.
However, these approaches have some shortcomings that need to be addressed.
(1) Lack of flexibility. 
Methods like LLaVA-PruMerge~\cite{shang2024llava-prumerge} employ identical vision tokens to represent an image for different questions in VQA. However, for datasets like MME~\cite{fu2023mme}, POPE~\cite{li2023evaluating_pope}, and AI2D~\cite{kembhavi2016ai2d}, numerous questions are associated with the same image, each focusing on a distinct area of the image, as illustrated in Figure\,\ref{fig:pope_example}(a).
Thus, LLaVA-PruMerge cannot dynamically adjust the important tokens based on the questions, leading to a lack of flexibility and a significant accuracy drop of $160.4$ on MME benchmark.
(2) Incomplete importance metric. 
Almost all methods necessitate the design of an importance metric for token reduction, such as cross-modal guidance for CrossGET~\cite{shi2023crossget} and attention score for FastV~\cite{chen2024fastv}.
However, there is limited theoretical evidence to establish which importance metric is the optimal.
A low importance score for a token does not necessarily indicate that the token is unimportant.
Moreover, some genuinely important tokens are pruned in these methods, as illustrated in Figure\,\ref{fig:pope_example}(b).
Thus, designing a comprehensive importance metric to determine which tokens are important is challenging.
(3) Incompatibility with KV Cache.
KV Cache~\cite{pope2023kv-cache} is an essential approach for speeding up MLLMs decoding by storing prior tokens' Key and Value states. 
FastV dynamically prunes partial vision tokens after a particular layer, with retained tokens varying for each auto-regressive prediction.
Thus, the cached KV of previous tokens cannot be reused because the retained vision tokens change with each prediction.
To use KV Cache to speed up the decoding stage, all KV Cache of vision tokens must be maintained in the initial prediction process. However, this leads to memory occupation for pruned tokens and reduces the memory benefit of token reduction. 
(4) Inconsistency with Flash-attention. FastV necessitates the retention of attention score for pruning  unimportant visual tokens effectively. Conversely, Flash-attention~\cite{dao2022flashattention}, despite its prevalent adoption, lacks the functionality to store the required attention scores for FastV. Since Flash-attention is used a lot, this limits how useful FastV can be.
Therefore, current methods don't fully meet the practical acceleration needs of MLLMs for various complex multimodal tasks.

\begin{figure}[!t]
  \centering
  \includegraphics[width=1\linewidth]{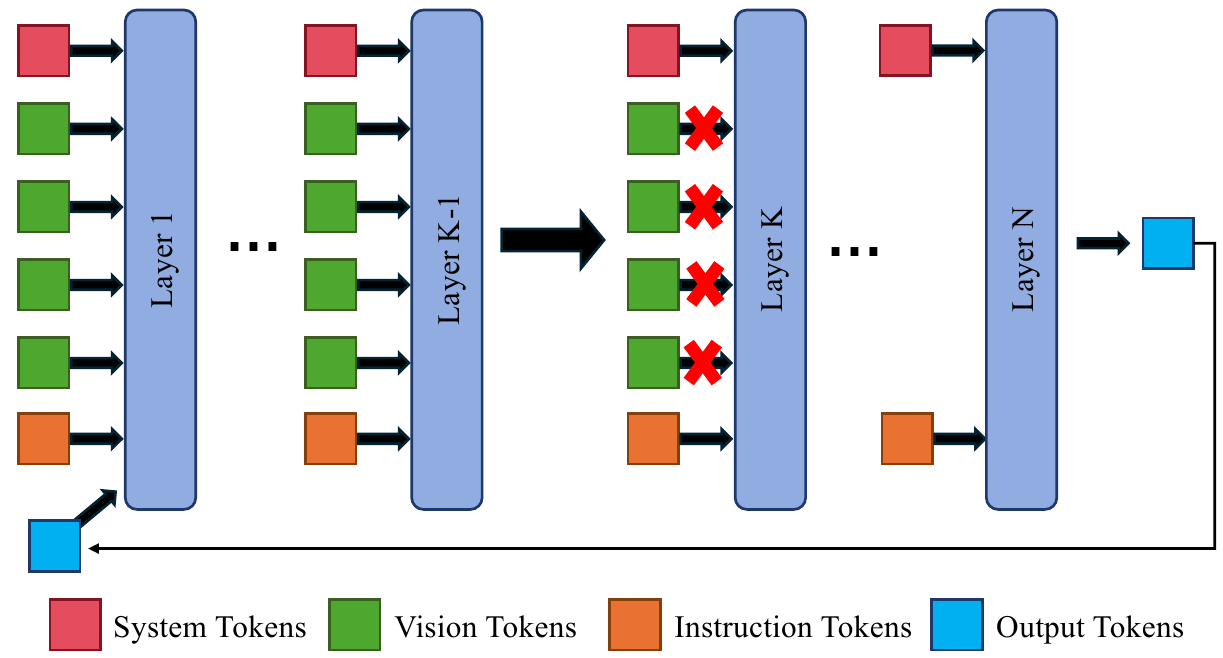}
  \caption{The framework of our method. Vision tokens are withdrawn in the $K$-th layer of  large language models.}
  \label{fig:framework}
  \vspace{-1.0em}
\end{figure}

Driven by the above analysis, we recognize that instead of being stuck in figuring out individual solutions to address the above shortcomings, a more comprehensive approach would be to withdraw all vision tokens after a specific layer.
In doing so, all vision tokens are preserved within the first few layers of MLLMs, ensuring flexibility across various multi-modal tasks.
Second, there is no need to design a comprehensive importance metric for selecting crucial tokens, thus avoiding the unintentional removal of important vision tokens.
Third, this approach is compatible with KV Cache because all vision tokens are preserved and removed simultaneously.
Finally, there is no need to store attention score for pruning unimportant vision tokens, thus this approach  aligns well with Flash-attention.
We conduct a thorough analysis to explore the possibility of removing all vision tokens after a specific layer.
In particular, we perform extensive visualizations and uncover some intriguing phenomena:
(1) The attention sink phenomenon in LLMs~\cite{xiao2023streamingllm} persists in MLLMs, where initial and nearest tokens gain the most attention, while middle vision tokens receive minimal attention.
% (e.g., $\sim$3\%)
%
(2) As the number of generated tokens increases, vision tokens receive less attention, while text tokens garner more attention. 
These results are attributed to the causal self-attention operation in LLMs, which only allows tokens to attend to preceding tokens, ensuring the model's generation depends on preceding content.
As a consequence of this operation, information from visual tokens migrates to subsequent text tokens through several layers of attention transformation.
Therefore, the latest token tends to pay progressively less attention to vision tokens and more attention to text tokens in the deeper layers and when there are more and more generated text tokens.

In this paper, we boost MLLMs with a Visual Tokens Withdrawal (VTW) strategy for rapid inference.
Given that vision tokens become less crucial in deep layers of LLMs and their information has already been absorbed by the subsequent text tokens, we propose withdrawing them in the deep layers.
VTM implements a visual tokens withdrawal approach at a specific layer of MLLMs, as illustrated in Figure\,\ref{fig:framework}.
Before this layer, computations proceed as usual; after this layer, vision tokens are removed, and only text tokens participate in the computation of deep layers.
To determine the withdrawal layer, we sample a small subset of datasets. 
Then, we calculate the Kullback-Leibler (KL) divergence between the output logits of the standard decoding and the visual tokens withdrawal decoding in each layer.
Finally, we select the first layer that meets the KL divergence criterion as the vision token withdrawal layer.

We carry out extensive experiments on various multimodal tasks, such as visual question answering ~\cite{kembhavi2016ai2d,lu2022sqa},  hallucination evaluation~\cite{li2023evaluating_pope}, visual reasoning~\cite{yue2023mmmu,fu2023mme} and video understanding~\cite{jang2017tgif,fu2024videomme} to show the efficacy of our VTW.
Notably, VTW can reduce over $40\%$ FLOPs on AI2D~\cite{kembhavi2016ai2d}, SQA\_image~\cite{lu2022sqa}, MMMU\_Val~\cite{yue2023mmmu}, MMB\_EN~\cite{Liu2023mmbench}, POPE~\cite{li2023evaluating_pope}, MME~\cite{fu2023mme}, TGIF~\cite{jang2017tgif}, and VideoMME~\cite{fu2024videomme} without compromising performance.
Also, VTW is applicable to the multimodal chatbot~\cite{2024llava} to achieve accelerated inference with  imperceptible differences in the answers.

% \clearpage

\section{Related Work}
\subsection{Multimodal Large Language Models}
Large Language Models (LLMs) like GPT~\cite{achiam2023gpt}, Gemini~\cite{team2023gemini}, and LLaMA~\cite{touvron2023llama,touvron2023llama2} have transformed natural language processing. They've been improved to understand not just text, but also images, video, audio,~\emph{etc}.
LLaVA~\cite{2024llava,liu2023llava1.5} combines a CLIP visual encoder~\cite{2021CLIP} with a LLaMA language decoder~\cite{2024llava}, making it good at following instructions and understanding images.
%
% BLIP-2~\cite{li2023blip2} connects an image encoder with a language model using a trainable Q-Former, achieving remarkable zero-shot image-to-text generation.
%
Video-LLaMA~\cite{lin2023video} endows videos and sounds understanding, improving how it processes different types of information.
However, these MLLMs use a lot of tokens to process information from different sources.
For example, LLaVA~\cite{2024llava} uses $576$ vision tokens for a $336\times336$ image, which escalates for higher-resolution images. 
This becomes a bigger issue in Video-LLaMA~\cite{lin2023video} and LLaVA-NeXT~\cite{liu2024llavanext}.
While MLLMs perform well, the high computational cost, which quadratically grows with a token number, is a big challenge.

\subsection{Vision Token Reduction}
Token pruning~\cite{rao2021dynamicvit,xu2022evo,liang2021evit} and merging~\cite{bolya2022tome,marin2021tokenpooling} directly reduce the number of tokens, thereby decreasing the inference time and memory usage.
EViT~\cite{liang2021evit} and Evo-ViT~\cite{xu2022evo} fuse non-critical tokens into a single token for token reduction.
ToMe~\cite{bolya2022tome} employs a binary soft-matching algorithm to merge redundant tokens, while Token Pooling~\cite{marin2021tokenpooling} utilizes clustering for token merging.
DiffRate~\cite{chen2023diffrate} and PPT~\cite{wu2023ppt} unify token pruning and merging to dynamically reduce redundant tokens.
In MLLMs, CrossGET~\cite{shi2023crossget} and MADTP~\cite{cao2024madtp} introduce special tokens to align tokens of different modalities and use these special tokens to guide token reduction.
Qwen-VL~\cite{Qwen-VL}, LLaVA-UHD~\cite{guo2025llava-uhd}, and LLaMA-VID~\cite{li2023llama-vid} use query tokens via cross attention to reduce vision tokens.
LLaVA-PruMerge~\cite{shang2024llava-prumerge} leverages the visional spatial redundancy and proposes a token reduction module that employs the similarity between the class token and spatial tokens as a key criterion for pruning and merging vision tokens.
It is observed that most image tokens receive inefficient attention after the second decoder layer~\cite{chen2024fastv}, thus half of the image tokens can be safely removed.

\section{Methodology}

\subsection{Motivations}
\label{sec:motivations}

MLLMs typically comprise a pre-trained vision encoder, a cross-modal projector, and a pre-trained large language model (LLM). Herein, we utilize LLaVA~\cite{2024llava}, a recent SOTA method, to illustrate the architecture.

The vision encoder, such as CLIP ViT-L~\cite{2021CLIP}, extracts visual features from an input image. These visual features are represented as a set of token sequences, referred to as vision tokens.
Then, a cross-modal projector transforms these vision tokens into the text embedding space, aligning the outputs of the vision encoder with the LLM.
The core is a pre-trained LLM, such as Vicuna~\cite{vicuna2023}, which is tasked with understanding the multimodal context and providing an appropriate response.
The integration of these components enables MLLMs to process and interpret both textual and visual data, providing a more comprehensive understanding of the inputs.

MLLMs require significant computational resources for inference, with the bulk of the computation being attributed to the LLM, given that the size of the visual encoder, such as  ViT-L (0.3B), is much smaller than the LLM, such as Vicuna (7B or 13B).
In MLLMs, the main computation for a decoder layer comes from multi-head attention (MHA) and feed-forward network (FFN). Assuming the input sequence length is $s$, the hidden embedding size is $h$, and the FFN up-scaling factor is $4$, the computational complexity for a transformer decoder layer is~\cite{chen2023cf-vit,chen2023diffrate,chen2024fastv}:
\begin{equation}\label{eq:inference_computation}
    \Omega(MHA+FFN) =2s^2h + 12sh^2,
\end{equation}
where computational complexity is quadratically influenced by the input length $s$. Therefore, efficient token management is crucial for optimizing the efficiency of MLLMs.

\begin{figure}[!t]
    \centering
    \includegraphics[width=\linewidth]{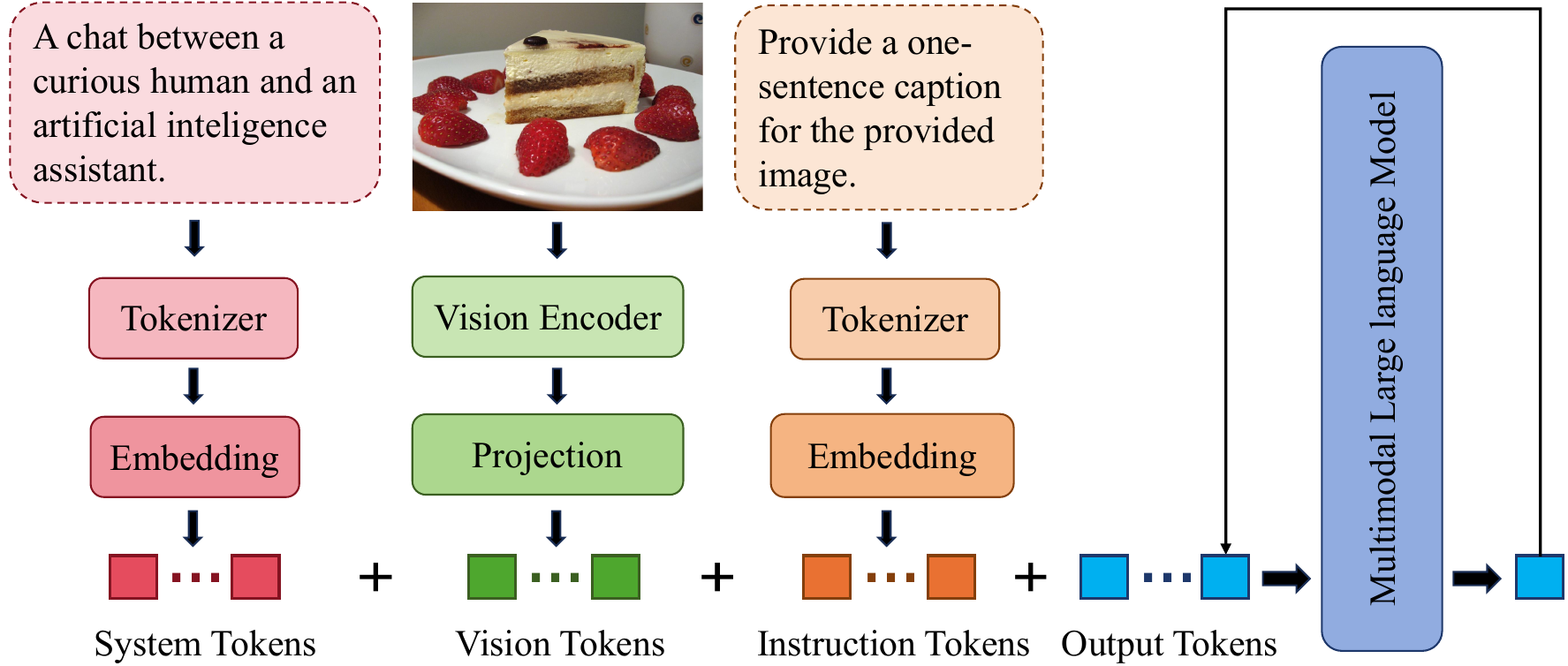}
    \vspace{-1.5em}
    \caption{The illustration for the input of a multimodal large language model. The input tokens are composed of system tokens, vision tokens, instruction tokens, and output tokens.}
    \label{fig:preliminary}
    \vspace{-1.5em}
\end{figure}

The input tokens of LLaVA in Figure\,\ref{fig:preliminary}, consist of system tokens, vision tokens, instruction tokens, and output tokens.
The system tokens are derived from fixed system prompts, which establish a dialogue system for LLaVA.
Meanwhile, the instruction tokens originate from users, which specify the query question for the given image.
In Figure\,\ref{fig:preliminary}, LLaVA preprocesses the input from various modalities and then concatenates all these tokens to form the inference input  as:
\begin{equation}\label{eq:llm_input}
X^t_1 = [S_1,\mathcal{V}_1,I_1,O_1^t], \\
\end{equation}
where $X_i^t$ is inputs of the $i$-th layer in the $t$-th inference. $S_1$, $\mathcal{V}_1$, and $I_1$ denote system tokens, vision tokens, and instruction tokens, with $O_1^t$ being the output tokens and $O_1^0 = \phi$. 
The quantity of instruction tokens varies depending on user instructions and is generally fewer than the number of vision tokens.
%
% The vision tokens account for the majority of the computation in the inference process, as they constitute most of the input tokens.
Vision tokens dominate the computation during inference by comprising the majority of input tokens.
Recent works~\cite{chen2024fastv,shang2024llava-prumerge} have attempted to reduce computation by decreasing the number of vision tokens.
However, these methods suffer from limited flexibility, incomplete importance metric, incompatibility with KV Cache, and inconsistency with the Flash-attention, as elaborated in the introduction.
\subsection{Unnecessity of Vision Tokens in Deep Layers of MLLMs}
\label{sec:analyze}

\begin{figure}[t]
    \centering
    \includegraphics[width=\linewidth]{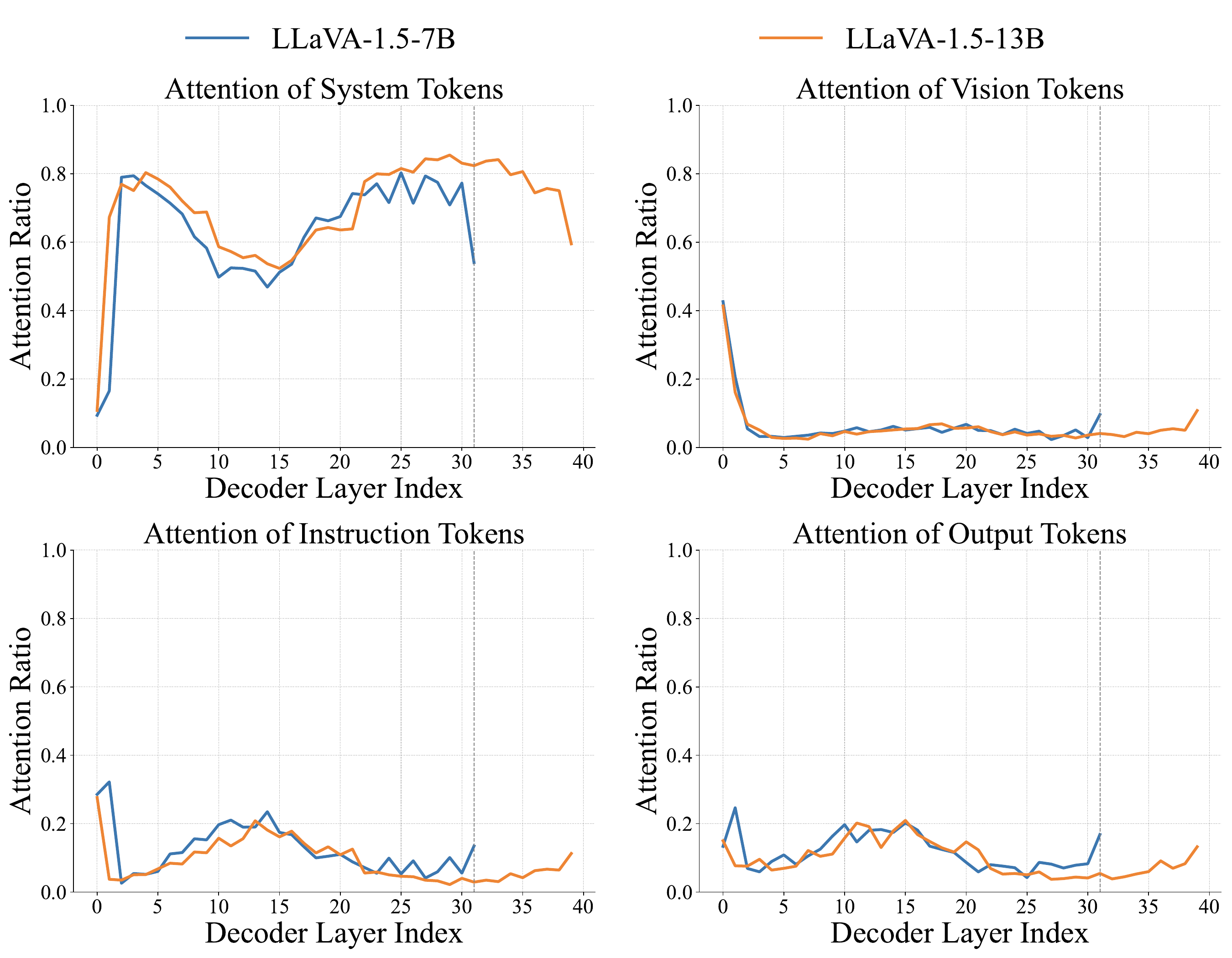}
\caption{The output token's attention towards various input token types across different layers on a combined subset of AI2D~\cite{kembhavi2016ai2d}, MMMU\_Val~\cite{yue2023mmmu}, MME~\cite{fu2023mme}, and POPE~\cite{pope2023kv-cache} (100 samples from each dataset).
The attention values are averaged across all attention heads and output tokens.
}
\vspace{-1em}
\label{fig:layer_attention}
\end{figure}

\begin{figure*}[t]
    \centering
    \includegraphics[width=\textwidth]{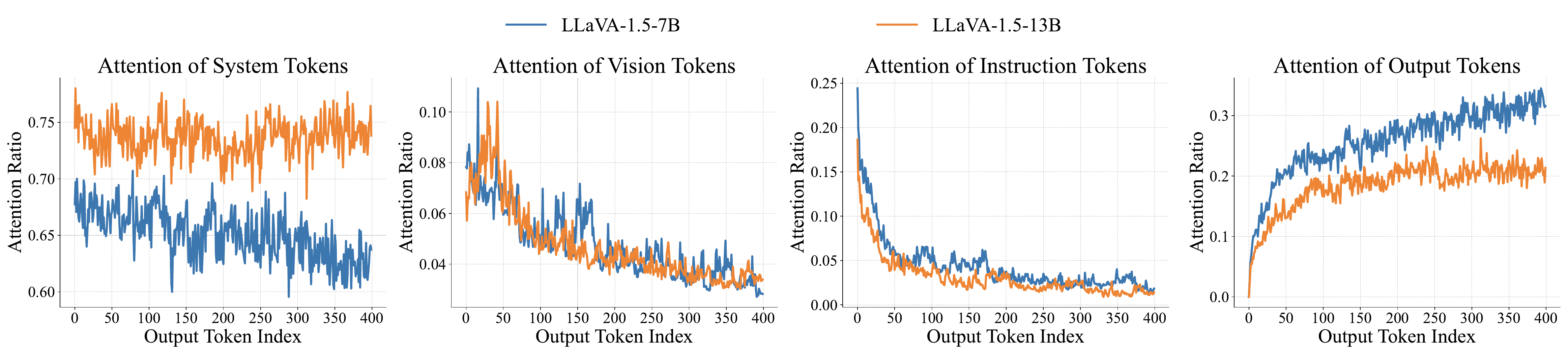}
    \caption{The output token's attention towards various input token types across output tokens. Our visualization is conducted on a subset of  AI2D~\cite{kembhavi2016ai2d}, MMMU\_Val~\cite{yue2023mmmu}, MME~\cite{fu2023mme}, and POPE~\cite{pope2023kv-cache} (20 samples from each dataset).
    The attention is averaged across all attention heads and layers.}
    \label{fig:output_attention}
    \vspace{-1em}
\end{figure*}

Given that input tokens $X^t_i$ consist of various types of tokens, it is natural to question  whether the contribution of each token type to the prediction of the output token is equal or proportional to their size.
Inspired by StreamingLLM~\cite{xiao2023streamingllm}, we opt for attention scores as the evaluation metrics.
Specifically, attention scores are derived from the causal self-attention operation within a decoder layer in LLMs.
In the $i$-th causal self-attention layer, the hidden feature $X^t_i$ is transformed into queries $Q_i^t$, keys $K_i^t$, and values $V^t_i$ using three distinct learnable projection matrices.
Consequently, the causal self-attention, abbreviated as CSA, can be expressed as:
\begin{equation}
    \text{CSA}(Q_i^t, K_i^t, V_i^t) = A_i^t \cdot V_i^t,
\end{equation}
where  attention map $A^t_i = \text{Softmax}(\frac{Q^t_i{K^t_i}^T + \Lambda }{\sqrt{d}})$ and $d$ is the hidden size of LLM.
$\Lambda$ is an upper triangular matrix where non-zero values are set to $-inf$ and diagonal elements are set to 0.
Here, for simplicity, we omit the expression of multi-head causal self-attention.
We select the last row of $A_i^t$ as the attention score $\alpha_i^t$, given that only the final token is utilized to predict the output token in an auto-regressive manner.
Based on the positions of system tokens, vision tokens, instruction tokens, and output tokens, we categorize $\alpha_i^t$ into $\alpha_i^{t, sys}$, $\alpha_i^{t, vis}$, $\alpha_i^{t, ins}$, and $\alpha_i^{t, out}$.
Then, we aggregate the attention scores of each input type, resulting $\beta^{t, sys}_i$, $\beta^{t, vis}_i$, $\beta^{t, ins}_i$, and $\beta^{t, out}_i$.
These variables denote the contributions of each input type to the output prediction.

We visualize $\beta$ towards various input token types across different layers in Figure\,\ref{fig:layer_attention} and across different output tokens in Figure\,\ref{fig:output_attention}. Here are the main points: 
% (1) In the initial layers of LLaVA, the model pays a lot more attention to system tokens and less to vision tokens.
(1) In the initial layers, attention to system tokens increases sharply, while attention to vision tokens decreases sharply.
In the middle and deeper layers, almost all the attention (80\% or more) goes to just 35 system tokens, while 576 vision tokens get only 5\%.
(2) As more output tokens are produced, the model focuses less on vision tokens and more on these output tokens.

The ``attention sink'' idea from LLM~\cite{xiao2023streamingllm} helps us understand how attention changes for system tokens in different layers. In the first few layers, the output token interacts with all other tokens through causal self-attention to accumulate semantic information. Consequently, the attention directed towards system tokens is minimal due to their limited semantic content. In the deep layers, the output token possesses sufficient self-contained information for its prediction. However, owing to the nature of the softmax function, which cannot assign zero attention to undesired tokens, the output token will mostly focus on tokens with minimal semantic information, such as system tokens, to avoid incorporating undesired information from other tokens.

\begin{table}[!t]
    \centering
    % \resizebox{1\linewidth}{!}{
    \begin{tabular}{ll}
    \toprule
       Experimental    Setting                          &   Score \\
       % Setting                                  &  Score      \\
    \midrule
        LLaVA-1.5-7B                                    &   1866.10  \\
    \midrule
      (a)   w/o image                                   &   970.89   \\
      (b)   w  non-content image at 1st--16-th layers     &   845.39   \\
      (c)  w original image at 1st--16-th layers          &   1872.43  \\
    \bottomrule
    \end{tabular}
    % }
    \caption{The ablation study  on MME~\cite{fu2023mme} under various experimental settings. In both (b) and (c), we remove vision tokens in the last 16-th layer of LLaVA-1.5-7b. ``non-content '' denotes a misleading image full of white area.}\label{tab:information migration}
\end{table}

To explain why instruction tokens and output tokens receive greater attention than vision tokens, we introduce ``information migration.''
Due to causal self-attention, tokens can attend only to preceding tokens. Instruction and output tokens can attend to all vision tokens, while vision tokens cannot attend to instruction tokens or output tokens.
After multiple transformations of causal self-attention layers, instruction and output tokens absorb both visual and textual information, attracting more attention from output tokens.

To verify that the information migration is completed within the first few layers of MLLMs, we design ablation studies using LLaVA-1.5-7B on MME benchmark.
As demonstrated in Table\,\ref{tab:information migration}, without any visual information, the score of LLaVA-1.5-7B decreases to $970.89$, emphasizing the significance of visual information in evaluation.
Interestingly, MLLMs can handle certain questions using only textual information, as evidenced by a total score of $970.89$.
By comparing the results of (a), (b), and (c), we observe that MLLMs achieve comparable results with baseline in the (c) setting  since the correct visual information has migrated to the subsequent text tokens before 16-th layer.
Besides, (b) yields a lower score than (a), as the misleading visual information has migrated to the subsequent text tokens  before 16-th layer.
These findings strongly support the occurrence of information migration within the initial layers of MLLMs.

Thus we can conclude that \textbf{the vision tokens are unnecessary in the deep layers of MLLMs}, as the information of vision tokens has migrated to following text tokens within the first few layers of MLLMs. Therefore, it is justifiable to withdraw all vision tokens in a specific layer of MLLMs.

\subsection{Visual Tokens Withdrawal} 
We review the standard inference process of MLLMs, followed by a comprehensive introduction to incorporate visual tokens withdrawal into the MLLMs framework.

Remember that the symbol $X_i^t$ represents the input of the $i$-th layer during the $t$-th inference pass of MLLMs, as in Eq.\,(\ref{eq:llm_input}).
The LLM employs an $N$-layer transformer architecture decoder to effectively predict output tokens:
\begin{equation}\label{eq:inferece}
\begin{aligned}
    X^t_{N+1} &= D_{1:N}(X^t_1),  \\ 
    X^{t+1}_1 &= [X^{t}_1, P(X^{t}_{N+1})],
\end{aligned}
\end{equation}
where $D_{i:j}(\cdot)$ denotes decoder layers from layer $i$ to layer $j$.
The $P(\cdot)$ predicts subsequent output tokens and calculates their embeddings by taking hidden features as input.

Keeping in mind that the vision tokens are not necessary in the deep layers of MLLMs due to the information migration, we proceed to withdraw these tokens at the $K$-th layer:
\begin{equation}\label{eq:vtw}
\begin{aligned}
    X^t_{K} &= D_{1:K-1}(X^t_1), \\ 
   \mathcal{X}^t_K &= X^t_K - \mathcal{V}^t_K, \\
   \mathcal{X}^t_{N+1} &= D_{K:N}(\mathcal{X}^t_K).
                   % &= [S_K,\mathcal{V}_K,I_K,O_K^t] - \mathcal{V}^t_K \\
                   % &= [S_K,I_K,O_K^t] \\
\end{aligned}
\end{equation}

In other words, before reaching the $K$-th layer, computations carry on as usual;
However, after the $K$-th layer, vision tokens are withdrawn, leaving only text tokens to engage in the computation of deep layers. Figure\,\ref{fig:framework} depicts an example.

Given the various variants and architectures of MLLMs, such as LLaVA-1.5-7B/13B~\cite{2024llava} and LLaVA-NeXT-7B~\cite{liu2024llavanext}, it is important to note that more complex tasks may require additional layers to process vision tokens for accurate predictions.
Consequently, the optimal withdrawal layer $K$ varies depending on the specific MLLMs and tasks at hand.
To determine the appropriate value for $K$, we randomly sample a tiny subset of the target datasets for guidance.
From Figure\,\ref{fig:layer_attention} that vision tokens receive minimal attention after layer $5$, we enumerate $K$ with values ranging from $5$ to $N$ and compute the KL divergence between the standard output logits  and the VTW's output logits. 
We then select the first layer that satisfies the  criterion as the withdrawal layer $K$:
\begin{equation}\label{eq:kl}
\begin{aligned} 
    KL(lm_{head}(X_{N+1}) ,lm_{head}(\mathcal{X}_{N+1}) ) < \eta, \\
    % \mathcal{X}_{N+1} &= D_{K:N}(\mathcal{X}^t_K) \\
\end{aligned}
\end{equation}
where, $KL(\cdot)$ shows KL divergence, $lm_{head}(\cdot)$ is the LLM project head, $X_{N+1}$ and $\mathcal{X}_{N+1}$ are calculated by Eq.\,(\ref{eq:inferece}) and Eq.\,(\ref{eq:vtw}) ,  and $\eta$ denotes the threshold.

\section{Experimentation}
\subsection{Experimental Setting}

VTW serves as a seamless extension to off-the-shelf pre-trained MLLMs, requiring no extra training cost.
We apply VTW to popular open-source MLLMs, such as LLaVA-1.5~\cite{2024llava}, LLaVA-NeXT~\cite{liu2024llavanext}, and Video-LLaVA~\cite{lin2023videollava}.
%
% LLaVA-NeXT achieves the best performance among other open-source MLLMs like CogVLM~\cite{wang2023cogvlm} and Yi-VL~\cite{young2024yi}.
%
We conduct comprehensive experiments across a multitude of multimodal tasks, such as VQA~\cite{kembhavi2016ai2d,lu2022sqa}, hallucination evaluation~\cite{li2023evaluating_pope}, visual reasoning~\cite{fu2023mme,yue2023mmmu,Liu2023mmbench}, and video understanding~\cite{fu2024videomme,jang2017tgif}.  
We use lmms-eval~\cite{Li2024lmms_eval} and VLMEvalKit~\cite{duan2024vlmevalkit} to evaluate MLLMs on different datasets.
%
% These experiments are to show our robustness and generalizability.
%
In our setup, the tiny subset size was set to 20, and the threshold $\eta$ was set to 0.003. 
For the multimodal chatbot, we withdraw vision tokens at the intermediate layers of MLLMs. 
% We withdraw vision tokens at the intermediate layers of multimodal chatbots. 

\begin{figure}[h]
    \centering
    \includegraphics[width=0.9\linewidth]{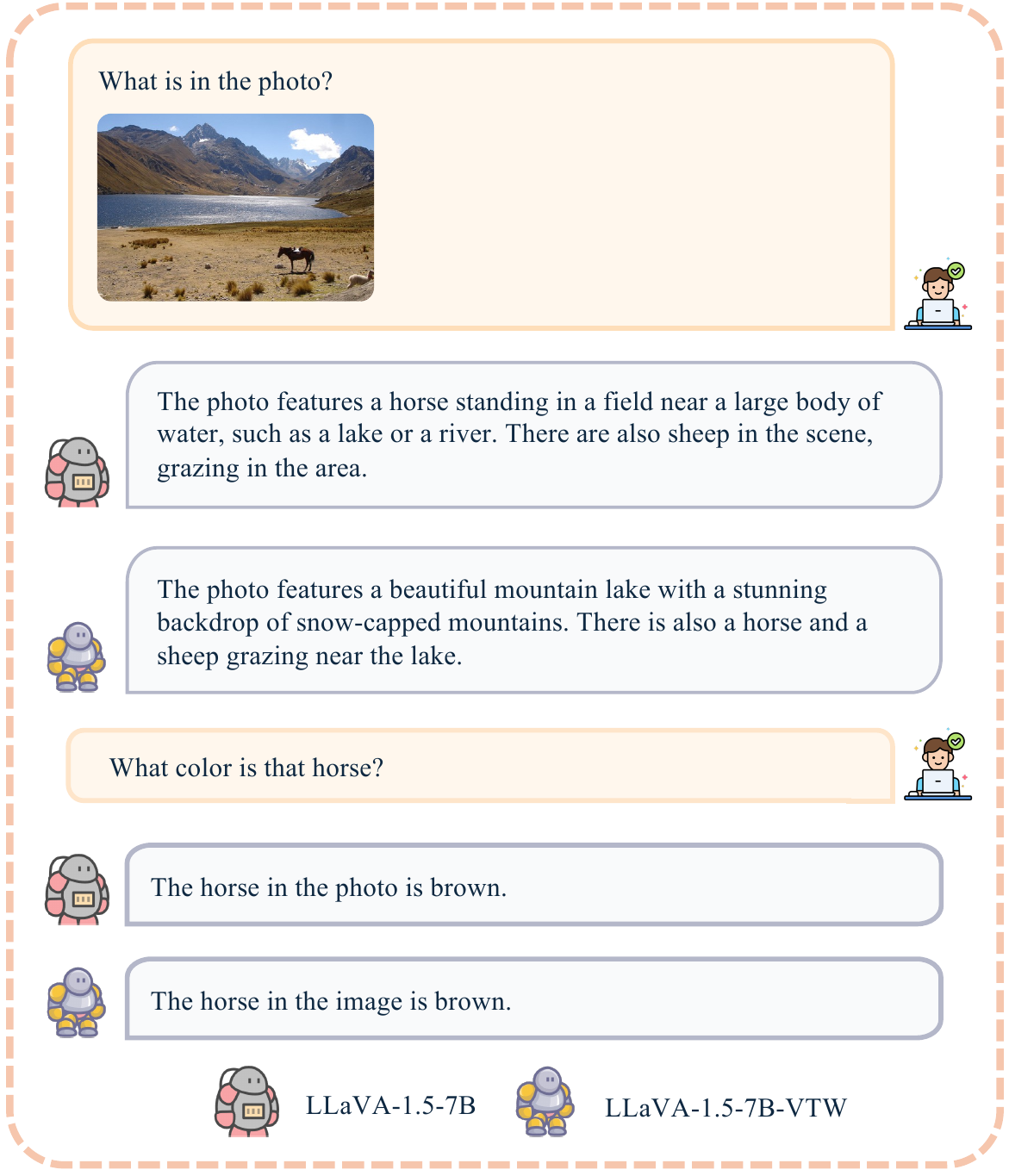}
    \caption{Dialogue comparison between baseline and VTW. }
    \label{fig:chatbot-7b}
    \vspace{-1em}
\end{figure}

\subsection{Multimodal Chatbot}

% We assess the effectiveness of our VTW in multimodal chatbots.
%
We aim to boost the chatbot's response speed without compromising response quality.
We compare the responses of MLLMs before and after applying VTW in  Figure\,\ref{fig:chatbot-7b}.
% \footnote{More results have been illustrated in supplementary material.}
%
We can see that the chatbots equipped with our VTW maintain their ability to produce the correct and similar outcomes compared to standard chatbots, despite lacking vision tokens in the deep layers.
This highlights VTW's success in accelerating the chatbot's responses while holding its performance.

\begin{table*}[!t]

    \centering
    % \resizebox{0.9\linewidth}{!}{
    \begin{tabular}{llccccccc}
    \toprule
          Methods            &  TFLOPs $\downarrow$ &AI2D $\uparrow$ & SQA\_Img $\uparrow$&  MMMU\_Val $\uparrow$&MMB\_{EN} $\uparrow$ &POPE $\uparrow$ & MME $\uparrow$\\
    \midrule
         LLaVA-1.5-7B        & 8.48 (100.00\%)      &   55.21        & 69.61                           & 35.60          & 64.09                              &   85.83        &   1866.10                       \\
         + FastV             & 4.91 (57.90\%)       &   55.14        & 68.96                           &  35.80         & \textbf{64.26}                     &   82.49        &   1864.35                        \\
         + Rand ($K$=16)       & 4.68 (55.19\%)       &    4.40        & 9.07                            &  29.30          & 0.52                        &   82.22           &   139.42                      \\
         + VTW$\dagger$ ($K$ = Random[8,24]) & $\approx$4.68 ($\approx$55.19\%)     &   55.36        & 69.21                           &  36.10 & 55.61                     &   76.05       &   1741.34                       \\
         + VTW ($K$=16)        &\textbf{4.68} (\textbf{55.19}\%)& \textbf{55.44} &\textbf{69.66}                   & \textbf{36.30} &   64.00                            & \textbf{85.96} &\textbf{1872.43}                     \\
    \midrule        
         LLaVA-1.5-13B       & 16.50 (100.00\%)     &  59.26         & 72.83                           & \textbf{34.90 } & 68.73                              &  \textbf{86.02}&    1827.26                            \\
         + FastV             & 9.56 (57.94\%)       & 58.87          & \textbf{73.03}                  &  34.60         &  68.30                             &  85.15         &\textbf{1855.11}                        \\
         + Rand ($K$=20)       & 9.10 (55.15\%)       & 2.49            & 7.93                         &  25.80         & 0.94                                &   83.78    &   41.74                    \\
         + VTW$\dagger$ ($K$ = Random[12,28]) & $\approx$9.10 ($\approx$55.15\%)     &  58.35      & 71.99   &  34.20          & 62.02                                  & 75.06    &   1661.36                   \\
         + VTW ($K$=20)        & \textbf{9.10} (\textbf{55.15}\%)& \textbf{59.39 }& 72.88                  & \textbf{34.90}&     \textbf{68.81}                 &   85.93        &    1828.79              \\
    \midrule
         LLaVA-NeXT-7B      & 28.73 (100.00\%)     & 65.31          & \textbf{70.15}                  &35.30           &   \textbf{67.18}                   & \textbf{86.44 }&  1846.33                         \\
         + FastV            & 15.67 (54.54\%)      & 64.86          & 68.96                           &  \textbf{35.70} &    66.84                          & 85.98          &  1786.17                            \\
         + Rand ($K$=16)      & 14.80 (51.51\%)      & 0.84      & 3.37                        &  24.30         & 00.17                                &   80.00          & 18.97                 \\
         + VTW$\dagger$ ($K$ = Random[8,24]) & $\approx$14.80 ($\approx$51.51\%)    & 64.54          & 69.86             &  35.40             & 59.28                          &   75.73          &  1723.65                    \\
         + VTW  ($K$=16)      &\textbf{14.80} (\textbf{51.51}\%) &\textbf{65.35 } & 70.00                        &  \textbf{35.70}         &  \textbf{ 67.18}                   &   86.33        & \textbf{1857.35}                 \\
    \bottomrule 
    \end{tabular}
    % }
    \caption{Comparison of various training-free methods for accelerating MLLMs inference. 
   SQA\_Img, MMMU\_Val, and MMB\_{EN} originate from ScienceQA~\cite{lu2022sqa}, MMMU~\cite{lu2022sqa}, and MMBench~\cite{Liu2023mmbench}, respectively.
   For a fair comparison, we manually set $K$ as 16 to keep VTW's FLOPs lower than FastV~\cite{chen2024fastv}. We use the average input length on MME to calculate TFLOPs. VTW$\dagger$ drops vision tokens in a random deep layer $K$. Rand randomly discards the same number of input
   tokens as the visual tokens. We employ \textbf{bold} formatting to highlight the best result. }    \label{tab:quantitative_results}
   \vspace{-1em}
\end{table*}

\subsection{Quantitative Evaluation}

\subsubsection{Visual Question Answering (VQA).}

In VQA, MLLMs interpret images before answering questions.
We test our VTW method on two popular VQA datasets: AI2D~\cite{kembhavi2016ai2d} and  SQA\_image~\cite{lu2022sqa}.
Table\,\ref{tab:quantitative_results} shows that VTW outperforms FastV and uses fewer FLOPs across different MLLMs.
It also achieves lossless acceleration compared to the baseline, with nearly half the FLOPs.
Notably, VTW outperforms the baseline by utilizing visual data in shallow layers and avoids excessive attention to irrelevant information that is considered noise.

\subsubsection{Visual Reasoning.}

Visual reasoning demands heightened perception, knowledge, and reasoning capabilities from the model compared to VQA.
We select MMMU\_Val~\cite{yue2023mmmu} and MMB\_EN~\cite{Liu2023mmbench} as our benchmarks for evaluation.
The results presented in Table\,\ref{tab:quantitative_results} demonstrate that VTW achieves comparable or even superior performance to the baseline and FastV, with fewer FLOPs.

\subsubsection{Hallucination Evaluation.}

Hallucinations can degrade MLLMs performance and severely impact user experiences in real-world applications.
We conduct experiments on POPE~\cite{li2023evaluating_pope} to investigate the impact of VTW on hallucinations.
As illustrated in Table\,\ref{tab:quantitative_results}, VTW achieves comparable performance to the baseline.
Conversely, FastV, which removes vision tokens after the second layer, exacerbates MLLMs' hallucinations, suggesting that premature removal of vision tokens exacerbates MLLMs hallucinations.

\subsubsection{Comprehensive Evaluation.}
MME~\cite{fu2023mme} accesses both perception and cognition across a total of 14 subtasks such as OCR, object localization, and attribute recognition.
% \footnote{We analyze the performance of each sub-task in supplementary material.} 
In Table\,\ref{tab:quantitative_results}, VTW demonstrates comparable performance to the baseline in a comprehensive benchmark, showing its excellent generalization ability in fine-grained tasks.

\subsubsection{Comparisons with Other VTW variants.}
In Table\,\ref{tab:quantitative_results}, Rand shows a significant drop in performance across all tasks, indicating that removing text tokens greatly impacts performance. 
VTW$\dagger$ shows that removing visual information too early affects performance.
VTW removes vision tokens in a proper layer, leading to the best performance.

\subsubsection{Video Understanding.}
We provide the results of VTW on different video question answering tasks (TGIF~\cite{jang2017tgif} and VideoMMe~\cite{fu2024videomme}) in Table\,\ref{tab:video_result}. 
VTW can generalize well in these video tasks and remain comparable to the baseline and FastV even with fewer FLOPs.

\begin{table}[!t]

    \centering
    % \resizebox{\columnwidth}{!}{
    \begin{tabular}{lrccc}
    \toprule
        \multirow{2}{*}{Model} & \multirow{2}{*}{FLOPs} & TGIF & VideoMME & \multirow{2}{*}{Avg} \\ 
        ~     &  ~    &  Acc  &  Overall& ~  \\
        \midrule
        Video-LLaVA          & 100.00\%                 & 0.25             & \textbf{0.30}  &  0.28 \\ 
        + FastV               & 53.13\%                  & 0.21            & \textbf{0.30}     &  0.26          \\
        + VTW ($K$=16)           & \textbf{50.00\%}         &\textbf{ 0.27}     & \textbf{0.30}   &   \textbf{0.29}       \\
        \bottomrule
    \end{tabular}
    % }
    \caption{Results on Video Question Answering Tasks. We only calculate vision tokens' FLOPs.}
    \label{tab:video_result}
    \vspace{-1em}
\end{table}

\begin{table}[t]
    \centering
    \begin{tabular}{lrrr}

    \toprule
         Metric            & Baseline   & FastV  &VTW\\ 
    \midrule
         KV Cache        &   \Checkmark     &    \Checkmark     &  \Checkmark         \\
          Flash Attention  &   \Checkmark     &   \XSolidBrush          &    \Checkmark       \\
    \midrule
        (a) Model GM                   & 14.1 G   & 14.1 G  & \textbf{14.1 G} (\textbf{1.00}$\times$)   \\
        (b) Peak Inference GM   & 17.1 G & 17.1  G                    & \textbf{16.1 G} (\textbf{0.94$\times$})  \\
        (b)-(a) Per Sample GM& 3.1 G & 3.1 G                                      & \textbf{2.0 G} (\textbf{0.65$\times$})  \\
    \midrule
        TFLOPs            & 22.9      & 12.9                            & \textbf{12.3} (\textbf{0.54$\times$})  \\
        Latency/Example   & 0.52 s    & 0.40 s                   & \textbf{0.33 s} (\textbf{0.63$\times$})  \\
    \bottomrule
    \end{tabular}
    \caption{The comparisons of practical GPU memory overhead and latency of LLaVa-NeXT-7B in SQA\_Image  on one NVIDIA RTX 3090. GM stands for GPU memory. }\label{tab:gpu_overhead}
\end{table}

\subsection{Cost Analysis}
The GPU memory overhead and latency comparisons between original MLLMs, FastV and VTW($K=16$) are presented in Table\,\ref{tab:gpu_overhead}.
We observe that during the inference of MLLMs, VTW reduces the GPU memory overhead by $35\%$ for each sample, while FastV needs to store all image tokens' KV Cache during the first forward, leading to a high peak GPU memory.
Furthermore, VTW reduces FLOPs by nearly half and the latency per sample to approximately $0.63 \times$ in comparison with the baseline models.

\begin{figure}[t]
    \centering
    \includegraphics[width=1\linewidth]{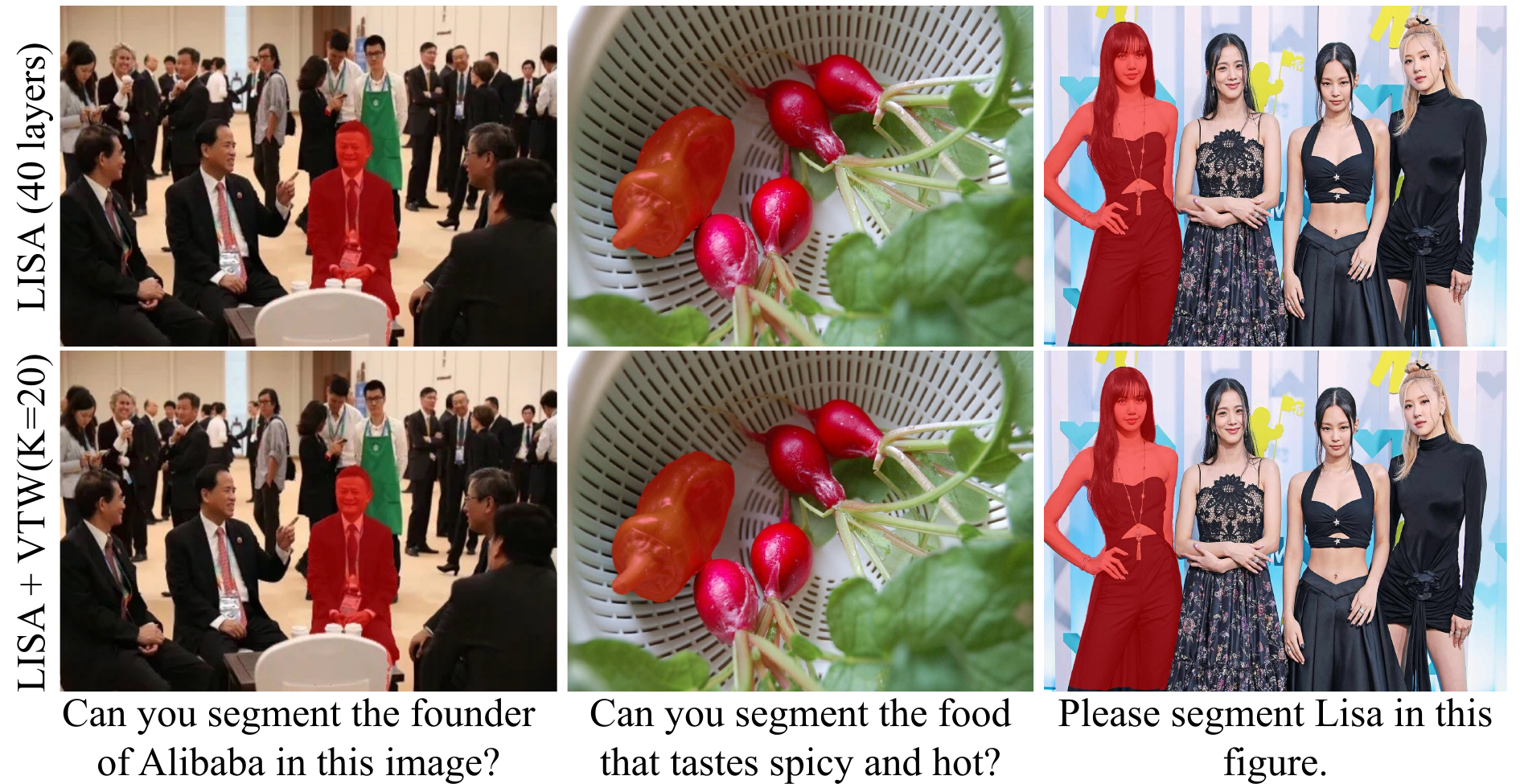}
    \caption{VTW's results on the segmentation task.}
    \label{fig:segment}
    \vspace{-1em}
\end{figure}

\subsection{Downstream Task}
To evaluate whether VTW is still workable in pixel-level fine-grained task, we apply it to LISA~\cite{lai2023lisa} which uses MLLMs for segmentation task. As visualized in Figure\,\ref{fig:segment}, VTW does not degrade the segmentation ability of LISA, showing its excellent generalization ability in the fine-grained downstream task.

\begin{figure}[!b]
    \centering
    \includegraphics[width=1\linewidth]{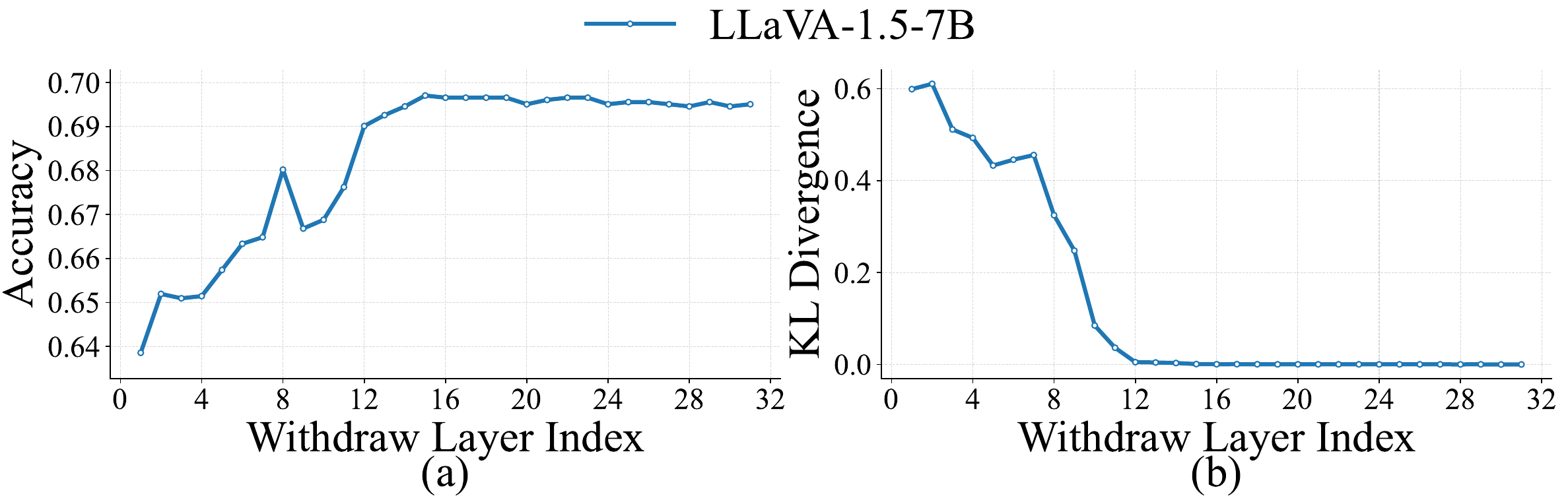}
    \caption{Ablation study on visual tokens withdrawal layer $K$. (a) Accuracy \emph{v.s.} K; (b) KL divergence \emph{v.s.} K.
    }
    \label{fig:ablation_k}
\end{figure}

\subsection{Ablation Studies}
We conduct ablations using LLaVA-1.5-7B on SQA\_Image.
\subsubsection{Ablations on Visual Tokens Withdrawal Layers.}

As depicted in Figure \ref{fig:ablation_k}(a), a small value of $K$, indicating an early withdrawal of vision tokens, leads to a degradation in the performance of MLLMs. 
When $K$ exceeds a specific layer number,  VTW performs similarly to the baseline. 
This observation further proves that vision tokens are unnecessary
in the deep layers of MLLMs.

\subsubsection{Ablations on Threshold $\eta$.}

Recall that we use KL divergence to determine the withdrawal layer $K$ on a tiny subset of target datasets, as formulated in Eq.\,(\ref{eq:kl}). 
As depicted in Figur\,\ref{fig:ablation_k}(b), the KL divergence  is large when $K$ is small, and it converges as $K$ exceeds a specific threshold. 
Combining the results from Figure\,\ref{fig:ablation_k} and Table\,\ref{tab:threshold}, we observe that when the KL divergence converges, the accuracy of VTW also converges to that of the baseline. 
Thus, we select KL as our criterion for determining the withdrawal layer $K$.

\begin{table}[t]

    \centering

    % \resizebox{\linewidth}{!}{
    \begin{tabular}{ccccccc}
    \toprule
        $\eta$ & 0.006  & 0.005  & 0.004 & 0.003  & 0.002  & 0.001   \\
    \midrule
         $K$       &   12  & 13   & 14   & 15   & 15  & 15  \\
    \midrule
        Accuracy &  69.01  & 69.26 &  69.46  & 69.70  & 69.70  & 69.70  \\
    \bottomrule
    \end{tabular}
    % }
    \caption{Ablation study on threshold $\eta$.}    \label{tab:threshold}
\end{table}

\begin{table}[!t]

    \centering
    % \resizebox{0.9\linewidth}{!}{
    \begin{tabular}{lccc}
    \toprule
         Position embedding       & SQA\_Image   & MMMU\_Val  & POPE\\
                            % & MME      & SQA   & MMMU  \\
    \midrule
         LLaVA-1.5-7B            & 69.61     & 35.60    &  85.83    \\
         Rearrange                 & 69.56     & 35.90   &  85.41    \\
         Keep                  & \textbf{69.66}& \textbf{36.30} &\textbf{85.96}\\
    \bottomrule
    \end{tabular}
    % }
    \caption{Ablation study on position embedding. ``Rearrange'' and ``Keep''  mean to rearrange  and keep the position embeddings  of  remaining  tokens after withdrawing vision tokens.}
    \label{tab:position embedding}
    \vspace{-1em}
\end{table}

\subsubsection{Ablations on  Position Embedding.}
LLMs use position embeddings (PE) to model the positional relationships between tokens. 
% 
% After withdrawing vision tokens, there are two ways to handle the PE of the remaining text tokens.
% % 
% One way is to keep the original PE, and the other is to rearrange the PE so that the text token positions are contiguous. 
After withdrawing vision tokens, the PE of the remaining text tokens can either be kept as is or rearranged for contiguous positions.
As shown in Table \ref{tab:position embedding},  “Keep”  outperforms  ``Rearrange'' in VTW across different datasets.

\subsection{Limitations and Future Works}
We are the first to find information migration happens in the initial layers of MLLMs. However, more complex tasks may need additional layers for this migration.
Thus, the reduction in FLOPs is relatively marginal for complex tasks compared to simpler ones.
Future works can try to apply VTW in the training stage to boost the information migration, which can save the training cost and further improve the performance of VTW.
There is also a chance to investigate the information migration in other modalities, like audio.

\section{Conclusion}
We have introduced visual tokens withdrawal (VTW), a plug-and-play module for faster MLLMs inference.
VTW is inspired by:
(1) The attention sink phenomenon in LLMs persists in MLLMs.
(2) The occurrence of information migration, indicates that visual information migrates to subsequent text tokens within the first few layers of MLLMs.
Building upon these observations, we deduce that vision tokens become unnecessary in the deep layers of MLLMs, despite they take up a significant computational overhead.
Thus, we withdraw vision tokens at specific layers in MLLMs.
Experiments across various multimodal tasks and chatbots validate the efficacy of VTW in boosting MLLMs for rapid inference without compromising performance.

\appendix
\section*{\centering \huge Appendix}
\section{More Multimodal Chatbot Results}

\begin{figure}[htbp]
    \centering
    \includegraphics[width=1\linewidth]{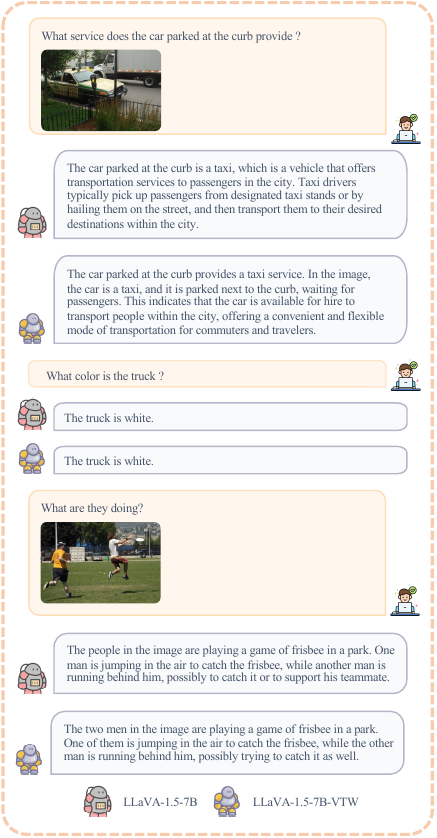}
    \caption{The chatbot dialogues comparison between the original LLaVA-1.5-7B and our LLaVA-1.5-7B-VTW. }
    \label{fig:chatbot-1}
\end{figure}

\begin{figure}[htbp]
    \centering
    \includegraphics[width=1\linewidth]{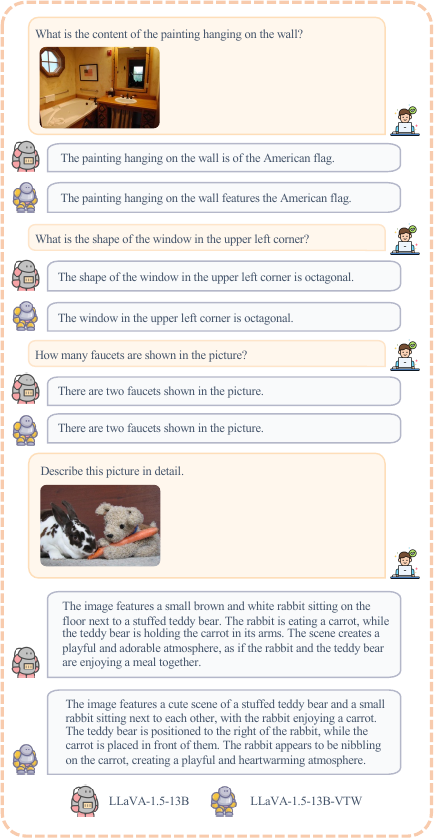}
    \caption{The chatbot dialogues comparison between the original LLaVA-1.5-13B and our LLaVA-1.5-13B-VTW. }
    \label{fig:chatbot-2}
\end{figure}

\begin{figure}[htbp]
    \centering
    \includegraphics[width=1\linewidth]{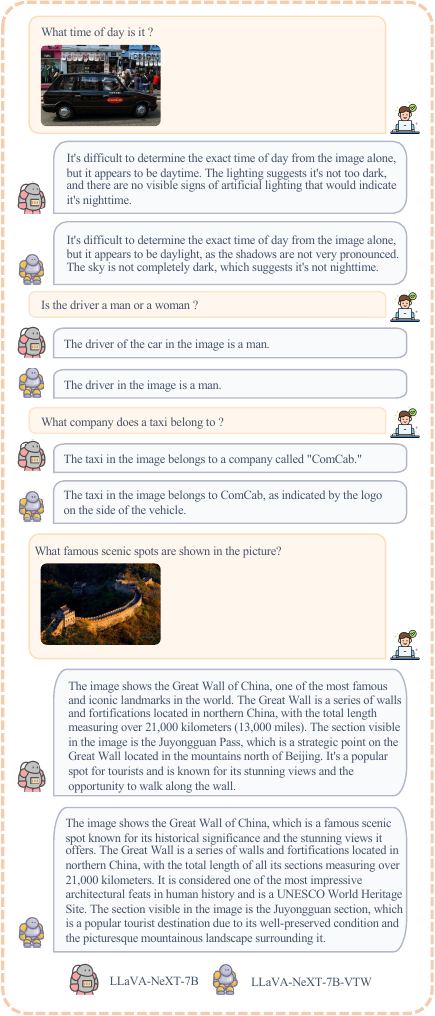}
    \caption{The chatbot dialogues comparison between the original LLaVA-NeXT-7B~\cite{liu2024llavanext} and our LLaVA-NeXT-7B-VTW. }
    \label{fig:chatbot-3}
\end{figure}

More multimodal chatbot dialogues are provided in Figure\,\ref{fig:chatbot-1}, Figure\,\ref{fig:chatbot-2}, and Figure\,\ref{fig:chatbot-3}.
We withdraw vision tokens at the 16-th, 16-th, and 20-th layers for LLaVA-1.5-7B~\cite{2024llava}, LLaVA-NeXT-7B~\cite{liu2024llavanext}, and LLaVA-1.5-13B~\cite{2024llava}, respectively.
Our VTW significantly reduces FLOPs by over 40\% for multimodal chatbots, while maintaining the quality of responses.
As shown in Figure\,\ref{fig:chatbot-1}, Figure\,\ref{fig:chatbot-2}, and Figure\,\ref{fig:chatbot-3}, the chatbots with our VTW are comparable to standard decoding in different multimodal tasks, such as VQA (answering the color of the truck and the gender of the driver), image caption (describing the detail of image), OCR (recognizing the logo of the car), and vision reasoning (reasoning the service of the car and the time of day).

\begin{table}[!ht]

    \centering
    % \setlength\tabcolsep{2pt}
    % \resizebox{0.8\linewidth}{!}{
    \begin{tabular}{lcc}
    \toprule
    Model  & gIoU & cIoU \\
    \midrule
    LISA-7B      & 0.88 & 0.94 \\
    +VTW $K$ = 16 & 0.88 & 0.94\\
    LISA-13B          &0.95 &0.97 \\
    +VTW $K$ = 20  &0.95 &0.97 \\
    \bottomrule
    \end{tabular}
    \caption{Quantitative results of VTW on downstream task. We test LISA with our VTW on the Reasonseg datasets.}
    % \vspace{-1em}
    % }
    \label{tab:Quantitative results}
\end{table}

\section{More Results on Downstream Task}
\begin{figure}[htbp]
    \centering
    \includegraphics[width=1\linewidth]{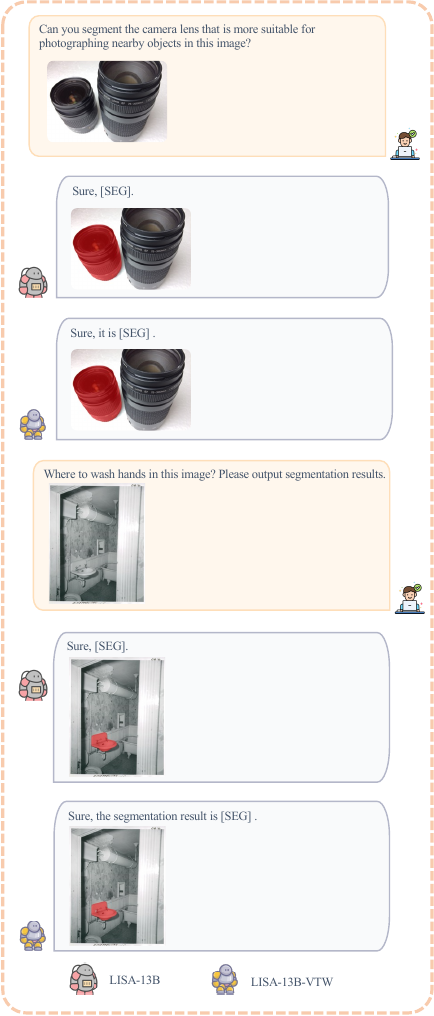}
    \caption{The chatbot dialogues and segmentation results comparisons between the original LISA-13B~\cite{lai2023lisa} and our LISA-13B-VTW.  }
    \label{fig:chatbot-lisa}
\end{figure}

LISA~\cite{lai2023lisa} inherits the language generation capabilities of the multi-modal large language model while also possessing the ability to produce segmentation masks.
We apply VTW on LISA to verify its effectiveness in fine-grained tasks, \emph{e.g.}, reasoning segmentation.
As shown in Figure\,\ref{fig:chatbot-lisa} and Table\,\ref{tab:Quantitative results}, VTW does not degrade the segmentation ability of LISA, showing its excellent generalization ability in the
fine-grained downstream task.

% \section{Experimental Environment}
% \begin{table}[!ht]

%     \centering
%     \setlength\tabcolsep{2pt}
%     % \resizebox{\columnwidth}{!}{
%     \begin{tabular}{cc}
%     \toprule
%     Hardware \& Software  & Specification \\
%     \midrule
%     Python      & 3.10 \\
%     Torch       & 2.1.2 \\
%     GPU         &  NVIDIA GeForce RTX 3090 \\
%     GPU Memory  & 24576M \\
%     Operating System  & Ubuntu 20.04.4 LTS \\
%     \bottomrule
%     \end{tabular}
%     \caption{The computing infrastructure used for  experiments.}
%     % }
%     \label{tab:setting}
% \end{table}

% We list the computing infrastructure and software used for  experiments in Table\,\ref{tab:setting}.

\section{Pseudo Code of Visual Tokens Withdrawal}

Algorithm\,\ref{alg:vtw} shows the process of Visual Tokens Withdrawal. In lines 1-9 of Algorithm\,\ref{alg:vtw}, we search for the appropriate visual tokens withdrawal layer k using a subset of the target dataset. Finally, we withdraw vision tokens in the searched layer k to boost multimodal large language models for rapid inference.

\begin{algorithm}[h]
    \small
    \SetAlgoLined
    \caption{\small VTW: Visual Tokens Withdrawal}
    \label{alg:vtw}
    \setcounter{AlgoLine}{0}
    \LinesNumbered
    
    \KwIn{Multimodal Large Language Models $\mathcal{M}$, target dataset $\mathcal{D}$, the total layers number $N$, threshold $\eta$ ;}
    \KwOut{Result $\mathcal{Y}$}

    Sample a subset of target datasets $\mathcal{D}$ as input $X$;
    
    Inference without VTW using Eq.(4) in main paper and get $X_{N}$;
    
    \For{$k\gets5$ \KwTo $N$} {

    Inference with VTW using Eq.(5) in main paper and get $\mathcal{X}_{N}$;
    
    Calculate $div$ between  $X_{N}$ and   $\mathcal{X}_{N}$ using Eq.(6);
    
    \If{$div < \eta$}{
    break
    }
    }

    Get results $\mathcal{Y}$ by evaluating $\mathcal{M}$ on $\mathcal{D}$ with VTW ($K=k$);
    
    \Return $\mathcal{Y}$
 
\end{algorithm}

\begin{table*}[!b] 
    \centering
    \resizebox{\linewidth}{!}{
    \begin{tabular}{lllllllllll}
    \toprule
          Methods            &  TFLOPs $\downarrow$ &AI2D $\uparrow$ & SQA\_Img $\uparrow$&  MMMU\_Val $\uparrow$&MMB\_{EN} $\uparrow$ &POPE $\uparrow$ & MME $\uparrow$ & Average  \\
    \midrule
         Qwen2-VL-2B-instruct (28 layers)& 100\% &   72.02        & 77.69                           & 36.00          & 71.90        &   87.86        &   1880.74             &         \\
         + VTW ($K$=18)        & 64\%     & 71.70 ($\downarrow$0.4\%) &77.94 ($\uparrow$0.3\%)& 37.00 ($\uparrow$2.7\%) &   70.03 ($\downarrow$2.6\%)  &  87.83($\downarrow$0.0\%) &1888.43 ($\uparrow$0.4\%)   & $\uparrow$0.1\%                \\
    \midrule        
        InternVL2-4B (32 layers) & 100\%     &  78.89        &96.33                          & 45.77  & 77.49                              &  84.60 &   2073.49           &              \\
         + VTW ($K$=16)       & 50\%     & 79.05 ($\uparrow$0.2\%) &96.23 ($\downarrow$0.1\%)& 45.22 ($\downarrow$1.2\%) &   77.49 ($\downarrow$0.0\%)  & 84.31(($\downarrow$0.3\%) & 2081.02 ($\uparrow$0.4\%)   & $\downarrow$0.2\%                 \\
    \bottomrule 
    \end{tabular}
    }
    \caption{Results of VTW on various MLLMs. SQA\_Img, MMMU\_Val, and MMB\_{EN} originate from ScienceQA~\cite{lu2022sqa}, MMMU~\cite{lu2022sqa}, and MMBench~\cite{Liu2023mmbench}, respectively. We compare the inference cost on vision tokens. }
    \label{tab:more mllms}
\end{table*}

\section{Results on More Kinds of MLLMs}
We present the results of VTW applied to various types of MLLMs in Table\,\ref{tab:more mllms}. Both Qwen-VL2~\cite{Qwen-VL} and Intern-VL2~\cite{chen2023internvl} are widely used, open-source multimodal large language models, each with an architecture that significantly differs from the LLaVA series~\cite{2024llava}. Our findings show that VTW reduces the inference cost for vision tokens by nearly 40\% with minimal performance degradation, underscoring its remarkable generalizability.

\begin{table*}[!t] 
    \centering    
    \resizebox{1\textwidth}{!}{
    \begin{tabular}{lccccccccccccccccc}
    \toprule
         Methods        & TFLOPs  & Existence & Count   & Position & Color & OCR & Poster & Celebrity  & Scene & Landmark & Artwork & Comm. & Num. & Text. & Code. & Total \\
    \midrule
         LLaVA-1.5-7B &8.48     & \textbf{190.00}       &155.00            &128.33              &170.00         &\textbf{140.00}   &\textbf{ 146.60}  &137.06             &158.00          &163.75           &119.50        &112.86     &\textbf{70.00}   &107.50   & \textbf{67.50}   & 1866.10\\
        + FastV       &4.91     & 180.00            &155.00            &\textbf{133.33}  &\textbf{180.00}  &137.50             &142.52           &\textbf{138.82}  &\textbf{160.00}   &\textbf{164.50}  &121.25         &116.43         &55.00  &\textbf{112.50} & 67.50  & 1864.35 \\
        + VTW (K=16)   &\textbf{4.68}    &\textbf{190.00}    &\textbf{160.00}   &128.33             &170.00          &137.50           &145.58              &135.59          &135.59         &163.75           &\textbf{123.25}   &\textbf{121.43}&65.00  &115.00 & 57.50  &\textbf{1872.43} \\
    \midrule
         LLaVA-1.5-13B&16.50    &\textbf{185.00}     &\textbf{155.00}    &\textbf{133.33}   &170.00 &125.00                     &\textbf{160.54}  & 152.06           &161.25        &\textbf{170.50 }  &\textbf{118.50}   &\textbf{128.57 } & \textbf{42.50 }&77.50 & 47.50  & 1827.26 \\
          + FastV     &9.56     &\textbf{185.00}     &\textbf{155.00 }   &\textbf{133.33 }  &\textbf{175.00} &\textbf{132.50 }  &159.52         &152.82          &\textbf{161.75 }  &168.25          &117.00            &126.43             &\textbf{42.50}  &\textbf{82.50 }  &\textbf{57.50 } & \textbf{1855.11}\\
          + VTW (K=16) &\textbf{9.10}     &\textbf{185.00}     &\textbf{155.00 }  &131.67   &\textbf{175.00} &125.00             &157.48         &\textbf{153.82 }    &161.25        &170.00          &\textbf{118.50 }&\textbf{128.57 }     & \textbf{42.50} &77.50 & 47.50  & 1828.79\\
    \midrule
         LLaVA-NeXT-7B&28.73  &\textbf{195.00}              &\textbf{135.00 }  &\textbf{143.33 }    &\textbf{170.00}  &\textbf{132.50} &\textbf{159.52}  &142.94            &162.25             &155.75  &123.25    &129.29                  &\textbf{42.50}  &100.00 & 55.00  & 1846.33\\
          + FastV    &15.67   &\textbf{195.00 }              & 125.00             &136.67   &\textbf{170.00 }         &117.50            &156.46             &140.29         &162.25             &156.75   &123.75    &\textbf{130.00}       &32.50            &87.50 &  52.50  & 1786.17 \\
          + VTW (K=16)&\textbf{14.80}  &\textbf{195.00}     &\textbf{135.00}   &133.33    &\textbf{170.00}          &\textbf{132.50}   &\textbf{159.52} &\textbf{144.71 }    &\textbf{163.00}  &\textbf{158.75  }  &\textbf{126.25 }  &129.29 &\textbf{42.50} &\textbf{105.00}  &\textbf{62.50}  & \textbf{1857.35}\\
    \bottomrule
    \end{tabular}
    }
    \caption{Comparison results for the subtasks of MME, including existence, count, position, color, OCR, poster, celebrity, scene, landmark, artwork, commonsense reasoning (Comm.), numerical calculation (Num.), text translation (Text.), and code reasoning (Code.). We utilize \textbf{bold} formatting to highlight the best result.}
    \label{tab:MME}
    \vspace{-1em}
\end{table*}

\section{Results on the MME's subtasks}
MME~\cite{fu2023mme} stands out as the first comprehensive benchmark for evaluating MLLMs, accessing both perception and cognition across a total of 14 subtasks, such as commonsense reasoning, numerical calculations, text translation, and more.
To thoroughly gauge the impact of VTW across different tasks, we present the scores for each task in Table\,\ref{tab:MME}.
VTW demonstrates comparable performance to the baseline across a majority of tasks, including existence, count, color, commonsense reasoning, and artwork. 
However, VTW exhibits marginal performance degradation on tasks such as position, poster, and OCR. 
These findings strongly suggest that the information encoded in vision tokens has migrated to subsequent text tokens within the first few layers of MLLMs.
To clarify, VTW only involves text tokens in the deep layers of MLLMs.
Despite the absence of vision tokens in the deep layers of multimodal large language model, VTW's performance remains on par with normal inference, where both vision and text tokens are used across all layers of MLLMs.

\clearpage

\section{Acknowledgments}
This work was supported by National Science and Technology Major Project (No. 2022ZD0118201), the National Science Fund for Distinguished Young Scholars (No.62025603), the National Natural Science Foundation of China (No. U21B2037, No. U22B2051, No. U23A20383, No. U21A20472, No. 62176222, No. 62176223, No. 62176226, No. 62072386, No. 62072387, No. 62072389, No. 62002305 and No. 62272401), and the Natural Science Foundation of Fujian Province of China (No. 2021J06003, No.2022J06001).

\bibliography{main}
\end{document}